\documentclass[lettersize,journal]{IEEEtran}
\usepackage{amsmath,amsfonts}
\usepackage{array}
\usepackage[caption=false,font=normalsize,labelfont=sf,textfont=sf]{subfig}
\usepackage{textcomp}
\usepackage{verbatim}
\usepackage{graphicx}
\usepackage{cite}
\usepackage{amssymb}
\usepackage{url}
\usepackage{hyperref}
\usepackage{glossaries}
\usepackage{adjustbox}
\usepackage[section]{placeins}
\usepackage{float}  
\usepackage{booktabs}
\usepackage{comment}

\newacronym{AI}{AI}{Artificial Intelligence}
\newacronym{GPT}{GPT}{Generative Pre-trained Transformer}
\newacronym{SOTA}{SOTA}{state-of-the-art}
\newacronym{CNN}{CNN}{Convolutional Neural Network}
\newacronym{VIT}{ViT}{Vision Transformer}
\newacronym{MLP}{MLP}{Multi-Layer Perceptron}
\newacronym{SSL}{SSL}{Self-Supervised Learning}
\newacronym{SL}{SL}{Supervised Learning}
\newacronym{MAE}{MAE}{Masked Autoencoder}
\newacronym{FCMAE}{FCMAE}{Fully Convolutional Masked Autoencoder}
\newacronym{GRN}{GRN}{Global Response Normalization}
\newacronym{ResMLP}{ResMLP}{Residual-based MLP}
\newacronym{HSW-MSA}{HSW-MSA}{Hybrid Shifted Windows Multi-Head Self-Attention}
\newacronym{Swin}{Swin}{Shifting Window}
\newacronym{KD}{KD}{Knowledge Distillation}
\newacronym{VQGAN}{VQGAN}{Vector Quantized Generative Adversarial Network}
\newacronym{GAN}{GAN}{Generative Adversarial Network}
\newacronym{PVT}{PVT}{Pyramid Vision Transformer}
\newacronym{BRA}{BRA}{Bi-level Routing Attention}
\newacronym{VAN}{VAN}{Visual Attention Network}
\newacronym{CLIP}{CLIP}{Contrastive Language-Image Pre-training}
\newacronym{VLM}{VLM}{Vision Language Model}
\newacronym{LM}{LM}{Language Model}
\newacronym{GRAFT}{GRAFT}{Ground Remote Alignment for Training}
\newacronym{FLOPS}{FLOPS}{Floating Point Operations per Second}
\newacronym{RSGPT}{RSGPT}{Remote Sensing Generative Pretrained Model}
\newacronym{MLLM}{MLLM}{Multimodal Large Language Model}
\newacronym{LLM}{LLM}{Large Language Model}
\newacronym{CCS}{CCS}{Cloud and Cloud Shadow}
\newacronym{TSSG}{TSSG}{Time-Series Spectral Similarity Group}
\newacronym{DOFA}{DOFA}{Dynamic One-For-All}
\newacronym{RS}{RS}{remote sensing}
\newacronym{MIM}{MIM}{Masked Image Modelling}
\newacronym{SI-SA GAN}{SI-SA GAN}{Spatial Information fusion Self-Attention Generative Adversarial Network}
\newacronym{SPAN}{SPAN}{Spatial Attention Network}
\newacronym{FM}{FM}{foundation model}
\newacronym{NIR}{NIR}{near-infrared}
\newacronym{SAR}{SAR}{Synthetic Aperture Radar}
\newacronym{GSD}{GSD}{Ground Sampling Distance}
\newacronym{FFN}{FFN}{Feed-Forward Network}
\newacronym{GFM}{GFM}{Geospatial Foundation Model}
\newacronym{CMID}{CMID}{Contrastive Mask Image Distillation}
\newacronym{FPN}{FPN}{Feature Pyramid Network}
\newacronym{ESA}{ESA}{European Space Agency}
\newacronym{NSO}{NSO}{Netherlands Space Office}
\newacronym{mIoU}{mIoU}{Mean Intersection Over Union}
\newacronym{mAP}{mAP}{Mean Average Precision}
\newacronym{DCNN}{DCNN}{Deep Convolutional Neural Network}
\newacronym{DDP}{DDP}{DistributedDataParallel}
\newacronym{MSE}{MSE}{Mean Squared Error}
\newacronym{CV}{CV}{Computer Vision}
\newacronym{POC}{POC}{Proof Of Concept}
\newacronym{MVP}{MVP}{Minimum Viable Product}
\newacronym{PCA}{PCA}{Principal Component Analysis}
\newacronym{tSNE}{tSNE}{t-distributed Stochastic Neighbor Embedding}
\newacronym{FDAF}{FDAF}{Feature Distance Attention Fusion}
\newacronym{AHN}{AHN}{Algemeen Hoogtebestand Nederland}
\newacronym{Grad-CAM}{Grad-CAM}{Gradient-weighted Class Activation Mapping}
\newacronym{PEFT}{PEFT}{Parameter-Efficient Fine-Tuning}
\makeglossaries
\begin{document}




\title{Developing a foundation model for high-resolution remote sensing
data of the Netherlands}

\author{
    Paul~Vermeeren, 
    Heysem~Kaya

\IEEEcompsocitemizethanks{
\IEEEcompsocthanksitem P. Vermeeren (paul@supato.nl) and H. Kaya (h.kaya@uu.nl) are with the Department of Information and Computer Sciences, Utrecht Univ., the Netherlands.
}
}  

\maketitle

\begin{abstract}
Foundation models are large-scale machine learning models trained on massive datasets, often in a self-supervised manner, allowing them to learn general representations that transfer well to many downstream tasks. They achieve strong performance on various downstream tasks while requiring less labeled data than models trained specifically for those tasks. This research aims to develop a foundation model using 1.2m high-resolution satellite images of the Netherlands. By combining a Convolutional Neural Network and a Vision Transformer, the model captures both low- and high-frequency landscape features, such as fine textures, edges, and small objects as well as large terrain structures, elevation patterns, and land-cover distributions. 
Leveraging temporal data as input, the model learns from broader contextual information across time, allowing the model to exploit the temporal dependencies, such as topographic features, land-cover changes, and seasonal dynamics. These additional constraints reduce feature ambiguity, improve representation learning, and enable better generalization with fewer labeled samples. The foundation model is evaluated on multiple downstream tasks, ranging from use cases within the Netherlands to global benchmarking datasets. On the vegetation monitoring dataset of the Netherlands, the model shows clear performance improvements by incorporating temporal information instead of relying on a single time point. Despite using a smaller model and less pretraining data—limited to the Netherlands—it achieves competitive results on global benchmarks when compared to state-of-the-art models. These results demonstrate that the model can learn rich, generalizable representations from limited data, achieving competitive performance on global benchmarks while using a fraction of the parameters of larger state-of-the-art remote sensing models. To maximize reproducibility and reuse, we made the scripts and the model accessible on \href{https://github.com/PaulVermeeren/Remote-sensing-foundation-model-for-the-Netherlands/}{GitHub}.

\end{abstract}





\section{Introduction}
\label{sec:intro}
\IEEEPARstart{I}{n} recent years, \gls{CV} has shifted from fully supervised, task-specific models to large-scale models that learn general visual representations. Traditional \gls{AI} methods rely on large labeled datasets, which are costly or infeasible for many domains, such as medical imaging, remote sensing, or long-term environmental monitoring \cite{szeliski2022computer, voulodimos2018deep}. Self-supervised learning addresses this limitation by exploiting the intrinsic structure in unlabeled data, requiring only a small number of labeled examples for downstream tasks \cite{khan2022transformers}. When trained at scale, these models become pretrained \glspl{FM} that can be adapted to a wide range of tasks with minimal supervision. \glspl{FM} have demonstrated strong performance in image understanding, object detection, and multimodal reasoning \cite{bommasani2022foundation}. \\

The development of \glspl{FM} enabled new applications that have significant societal and environmental benefits. For example, models can help with floods or forest wildfire burn scars, not only identifying current damage, but also learning from historical events to understand their causes, progression, and effects \cite{Prithvi-100M-preprint}. By analyzing sequences of images over time, these models can detect emerging patterns, predict future events, and provide early warnings, which would be difficult to do from single snapshots. Other applications enabled by \glspl{FM} include monitoring animal populations, tracking vegetation growth, and monitoring effects of climate change, which is almost impossible to track at large scale only from observations on the ground. \\

In this paper, we propose a novel method for training an \gls{FM} on high-resolution \gls{RS} imagery of the Netherlands. Unlike globally trained \glspl{FM}, which are optimized for diverse but lower-resolution datasets, a model trained on a country-specific high-resolution dataset can capture fine-grained spatial patterns, local land-use structures, and seasonal dynamics that are otherwise underrepresented. Most vision-based \glspl{FM} are trained in static imagery, although many real-world applications contain temporal aspects. This work explores how temporal signals can be used in an \gls{FM} to improve the learning of \gls{RS} imagery representation. The temporal relations provide additional context, such as seasonal and event-based changes, allowing the model to learn more robust and physically consistent representations. \\

High-resolution temporal \gls{RS} imagery contains complex multi-scale spatial patterns and temporal dynamics. While CNNs capture local fine-scale features effectively, they struggle with long-range spatial dependencies. Transformers, by contrast, excel at capturing global patterns through self-attention. To combine these strengths, this work develops a hybrid \gls{SOTA} architecture that efficiently extracts both local details and global context, such as individual objects such as buildings, roads, and vegetation patches, as well as large-scale patterns including urban regions, agricultural areas, and river networks. By specifically pretraining an \gls{FM} on high-resolution data, the goal is to bridge the gap between high-resolution \gls{RS} imagery and \glspl{FM}, allowing more accurate, locally adapted representations that outperform globally trained models on region-specific tasks. Data from the \gls{NSO} was used which is sourced from the Pléiades NEO and SuperView NEO satellites. \\

The remainder of this paper is organized as follows. Section~\ref{sec:literature} reviews relevant algorithms for general computer vision and remote sensing, including preprocessing, architectures, and fine-tuning for downstream tasks, as well as methods for assessing the environmental impact of foundation models. Section~\ref{sec:methodology} presents the methodology, detailing the proposed model design, considerations for computational complexity, performance, versatility, and environmental impact, and preliminary experiments for fine-tuning. Section~\ref{sec:results} presents experimental results, benchmarking the model against other \gls{SOTA} \gls{RS} models. Section~\ref{sec:conclusion} concludes and outlines future directions.

\section{Related Work}
\label{sec:literature}

Adapting general vision models to specific \gls{RS} tasks is more complex than re-training on new data: the differing \gls{GSD}, spatial–temporal characteristics and the need for multi-spectral analysis demand tailored architectures. This section therefore focuses on recently adapted models for \gls{RS}, \glspl{VIT} and hybrid \gls{VIT}–\gls{CNN} networks. Most studies in this section share some of the same benchmarks: EuroSAT, RESISC-45, BigEarthNet and UC-Merced for scene classification and OSCD for object and change detection. These will be used to compare the different architectures.

\subsection{Vision Transformers for Remote Sensing}

Recent work has adapted ViT architectures to the specific traits of remote sensing imagery by integrating spectral, temporal and spatial features while controlling computational cost. SatlasNet implements early multiscale temporal modeling, which is based on a hierarchical Swin Transformer to process multispectral time series, producing feature pyramids that feed dedicated decoder heads for diverse tasks from land cover mapping to change detection \cite{bastani2023satlaspretrain}. A similar multi-spectral multi-scale training and reconstruction approach is built using a \gls{MAE} in SatMAE++ \cite{noman2024rethinking}. The architecture is similar to \gls{MAE}, but the input image is downsampled twice before being masked and parsed to the encoder. The decoder reconstructs two upsampled scales to the original resolution, requiring the encoder to represent smaller scales effectively. \\

Building on these multi-scale, multi-spectral reconstructions, Prithvi-100M extends the MAE paradigm into three dimensions, width, height, and time, by initializing its encoder from ViT-L and training on long sequences to capture evolving spatial patterns and deliver robust temporal embeddings \cite{Prithvi-100M-preprint}. To unify spectral modalities, DOFA introduces a hypernetwork that generates wavelength-specific convolutional filters within a shared MAE backbone, dynamically adapting to optical, multi-spectral, and SAR bands and reaching new state-of-the-art performance across 12 of 13 benchmarks \cite{xiong2024neural}. Finally, SpectralGPT merges all channels into high-density 3D tokens under a 90\% masking scheme, learning joint spectral-spatial positional embeddings that enable high-fidelity reconstruction of both spectral signatures and spatial structures with minimal supervision \cite{hong2023spectralgpt}.

\subsection{Hybrid Vision Transformer - Convolutional Neural Network for
Remote Sensing}

Researchers have shown that using \glspl{CNN} with \glspl{VIT} can limit the cost of the model while improving both local texture sensitivity and global context awareness. The MATTER model learns material and texture representations by contrasting anchor, positive, and negative samples~ \cite{akiva2022self}. It uses a pixel-adaptive convolutional kernel to focus on fine-grained patterns through cosine similarity and a surface residual encoder to compute the distance of each anchor from learned texture groups, avoiding mask-based occlusion techniques and directly targeting material recognition instead. Building on this blend of locality and globality, CTAM augments the YOLOv8n backbone with a lightweight transformer attention module: feature maps at four scales are combined element-wise with transformer outputs to introduce global context without sacrificing real-time performance \cite{lang2024convolution}. 

RingMo-lite extends this hybrid approach by processing each image patch in parallel with a shallow CNN and a Swin transformer. Its PIMask strategy, unmasking smaller subpatches within masked regions, reduces data requirements to train the Swin \gls{MAE} \cite{wang2023ringmo}. As Table \ref{tab:performance_table} shows, these hybrid architectures achieve competitive top-1 accuracy, mAP, and F1 scores across diverse remote sensing benchmarks, demonstrating that combining convolutional layers with transformer attention remains a promising path to efficient high-performance models.

\begin{table*}
\centering
\caption[Performance evaluation of different models on various remote sensing datasets.]{Performance evaluation of different models on various remote sensing datasets, using top-1 accuracy, mean average precision (mAP), and F1 scores. The parameters and FLOPS are listed for the models and datasets it is available for in the literature. For BigEarthNet, only 10\% of the training data is used for fine-tuning.}
\label{tab:performance_table}
\begin{adjustbox}{width=\textwidth}
\begin{tabular}{llcccccc}
\toprule
& & &\multicolumn{1}{c}{EuroSAT \cite{helber2019eurosat}} & \multicolumn{1}{c}{RESISC-45 \cite{resisc45}} & \multicolumn{1}{c}{BigEarthNet \cite{bigearthnet}} & \multicolumn{1}{c}{UC-merced \cite{ucmerced}} & \multicolumn{1}{c}{OSCD \cite{oscd}}  \\
\midrule
Task & & & Classification & Classification & Classification & Classification & Change detection \\
\midrule 
Architecture & Backbone & & Top-1 & Top-1 & mAP 10\% & Top-1 & F1 \\ \midrule
\multicolumn{8}{l}{\textbf{Vision Transformer}} \\
SpectralGPT \cite{hong2023spectralgpt} &ViT-B  &  & \textbf{99.21} &- & \textbf{88.22} & -& \textbf{54.29} \\
SatMAE++ \cite{noman2024rethinking} &ViT-L & & &99.04 & \textbf{97.48} & 85.11 & 97.62 \\
DOFA \cite{xiong2024neural} &ViT-L & & 93.80 & 96.10 & - & - & -  \\
\multicolumn{8}{l}{\textbf{Hybrid ViT-CNN}} \\
MATTER \cite{akiva2022self} &ResNet-34  &  &- & 87.98 & -&-& 49.48 \\
RingMo-lite \cite{bastani2023satlaspretrain} &ImageNet-22k & &  -& 93.25 & - & \textbf{99.05} & -  \\
\bottomrule
\end{tabular}
\end{adjustbox}

\end{table*}

\subsection{Downstream Tasks for Remote Sensing}

\glspl{FM} provide a versatile backbone for diverse downstream tasks: once the encoder’s layers are frozen, the rich embeddings they produce can be fed into decoder heads tailored to segmentation, detection or classification. In the object detection and land‐cover classification framework of Zhang et al. \cite{zhang2022consecutive}, outputs from multiple transformer blocks are aggregated by a Feature Pyramid Network (FPN) to capture both fine and coarse spatial details. Land‐cover classification employs UPerNet \cite{xiao2018unified} to merge multi‐scale features into a pixel‐wise map of surface types. Scene classification, by contrast, simply applies global average pooling to the final transformer features and uses a linear layer to predict the overall scene category. 

RSBuilding \cite{wang2024rsbuilding} extends this paradigm to change detection: it feeds “before” and “after” image pairs through ViT and Swin encoders, then uses a cross‐attention decoder with learnable prompts that guide focus toward newly constructed buildings, with each decoder layer learning task‐specific embeddings. SegFormer~\cite{xie2021segformer} takes a U-Net–style approach within a transformer framework: each hierarchical encoder stage is passed through an MLP decoder block—first merging spectral dimensions, then spatial dimensions—and the final merged features are linearly projected to produce the segmentation mask.

\section{Methodology}
\label{sec:methodology}

This section outlines the methodology adopted in this work, including data acquisition, preprocessing steps, and the design of the network architecture. The consideration for the choice of data sources, the characteristics of the dataset, as well as the preprocessing techniques applied, are detailed in Subsection \ref{sub:data}. To conclude this section, Subsection \ref{sub:architecture} presents the architecture developed for this study.

\subsection{Data Acquisition and Preprocessing} \label{sub:data}
Determining the optimal combination of resolution, channels, image size, and quantities to train an \gls{FM} is essential to maximize model performance and computational efficiency. The \gls{NSO} offers the Pléiades Neo and Superview Neo satellites, which together cover the Netherlands 6 times yearly. Both satellites offer the optical bands red, green, blue and narrow IR, with Pléiades Neo offering an additional red edge and deep blue. The bands have a \gls{GSD} of 1.2m and are pan-sharpened to 0.3m. An applied preprocessing step is padding two channels for the Superview Neo images, while these only contain four bands, whereas Pléiades Neo contains six bands. These padded channels do not contribute to the gradient and solely act as a placeholder. The created pre-training dataset resulted in a total of 52,222 data points, each consisting of six images selected between March 16, 2023, and March 15, 2024. This resulted in a dataset that covers all four seasons to learn the temporal and seasonal variation. The complete dataset consists of 310,000 images and is split into a train, validation, and test set using a 70/20/10 ratio, resulting in 220,000, 60,000, and 30,000 images, respectively. The different splits cover separate areas, and each sample consists of six timestamps that capture the same location at different times. These images are cut into 512x512px tiles to be universally processed by the model. \\

\begin{figure}[h]
    \centering
    \includegraphics[width=0.95\columnwidth]{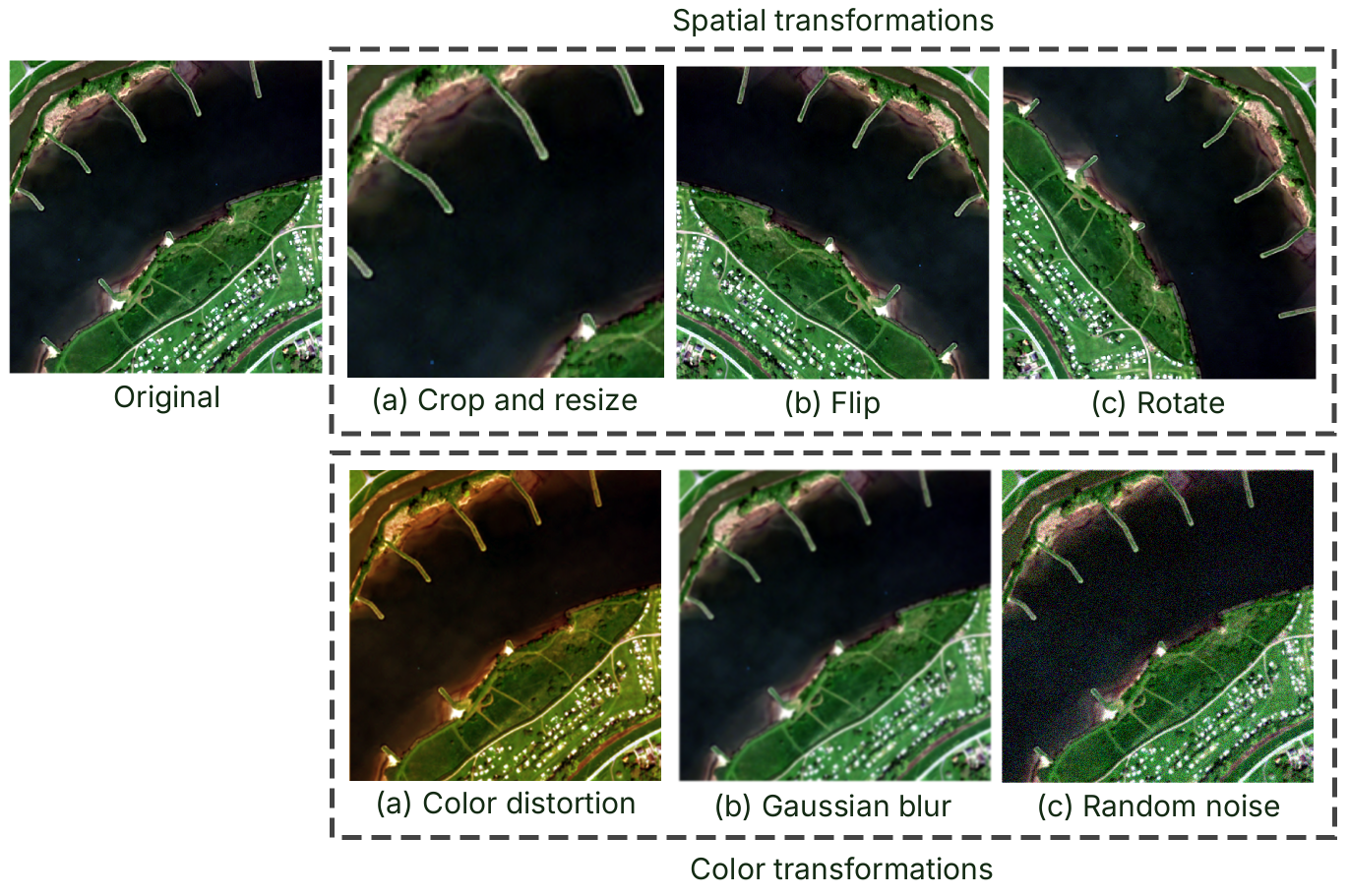}
    \caption[Illustration of the different augmentation transformations.]{Illustration of the different augmentation transformations grouped as spatial transformations and simulating temporal transformations.}
    \label{fig:augmentations}
\end{figure}

The model learns from multiple observations of the same location by processing 6 timestamps at once. As stated in \cite{cong2022satmae}, the incorporation of repeated multi-spectral observations provides the model with a richer context. This enables the model to learn characteristics under various real-world conditions such as seasonal and weather-related changes. To reduce overfitting and improve learning, data augmentation is applied by duplicating samples with spatial and/or color transformations. Spatial transformations include operations such as random cropping, flipping, and rotating, while color transformations include color distortion, Gaussian blur, and adding random noise. Spatial transformations mimic variations observed in images captured at different geographic locations, whereas color transformations simulate changes that occur in the different observations, such as variations in lighting due to weather \cite{li2021remote}. To preserve temporal consistency, all augmentations are applied equally across the temporal image sequence used to train the \gls{FM}. Figure \ref{fig:augmentations} illustrates the applied augmentations.

\subsection{Architecture} \label{sub:architecture}

\begin{figure*}[htbp]
    \centering
    \makebox[\textwidth][c]{\includegraphics[width=\linewidth]{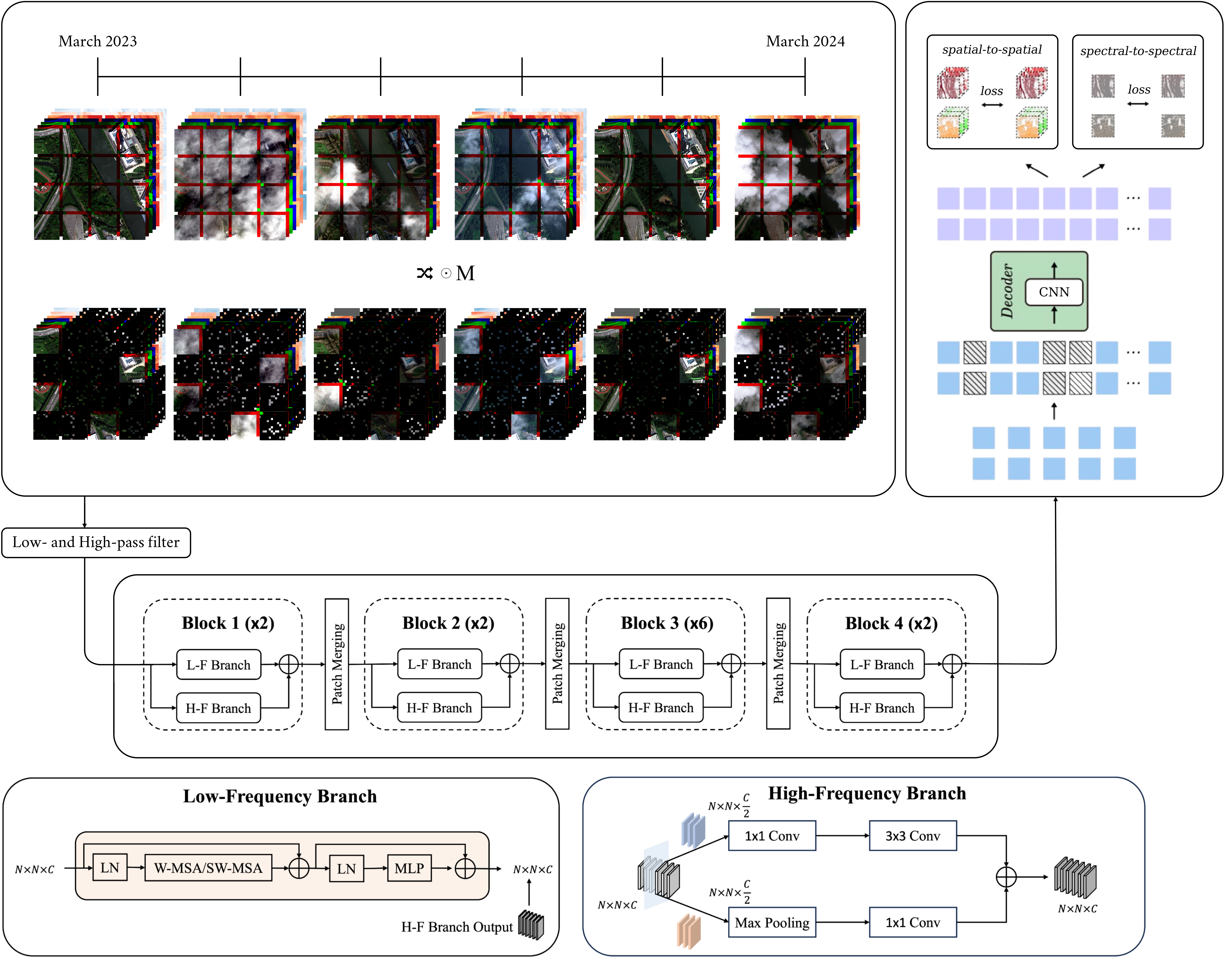}}
    \caption[The architecture for the pretraining the model.]{The composed architecture for the pretraining the model including a 4D patch embedding inspired by SpectralGPT \cite{hong2023spectralgpt}, and four patch merging steps using a Low-Frequency (L-F) Branch for local features and High-Frequency (H-F) Branch for global features inspired by RingMo-Lite \cite{wang2023ringmo}. These branches use a CNN and a \gls{VIT} architecture, respectively, to leverage the advantages of each method.}
    \label{fig:proposed_architecture}
\end{figure*}

The architecture of \gls{FM} is visualized in Figure \ref{fig:proposed_architecture} and is designed to limit computational complexity while maximizing performance. This architecture integrates a \gls{VIT}-\gls{CNN} structure, leveraging both methods strengths, inspired by RingMo-Lite \cite{wang2023ringmo}. This paper used the Swin Transformer-integrated \gls{MAE} framework as a starting point \cite{dai2023swin}, extending it with a CNN-based branch and a loss focussing on spatial and spectral properties of the data. \\

\subsubsection{Self-supervised pretraining}
Using \gls{MIM}, the model learns to create self-supervised rich embeddings to reconstruct the original image. Instead of traditional masking as in the vanilla \gls{MAE}, the 3D tensor masking approach as proposed in SpectralGPT is adapted to handle the spatial-spectral-temporal data \cite{hong2023spectralgpt}. This approach of processing a series of spectral images is similar to SatlasNet \cite{bastani2023satlaspretrain} and Prithvi-100M \cite{Prithvi-100M-preprint}, improving spatial and temporal feature detection. Examples are objects that change throughout the year and being able to detect the same object at different moments.  \\

\subsubsection{Spatiotemporal input representation}
The 3D tensor approach of SpectralGPT \cite{hong2023spectralgpt}, including width, height, and spectral channels, is improved by adding a fourth dimension of time. Given the 4D space containing a sequence of 3D spectral images  $\mathbf{x} \in \mathbb{R}^{H \times W \times C \times T}$, each image is converted into 3D patches. These patches have size $p \times p \times C$, where $p$ and $C$ are the size of the spatial and spectral dimensions of the patch sizes respectively. This results in $\frac{H}{p} \times \frac{W}{p} \times C \times T$ patches, which comes down to $\mathbf{x} = \{ \mathbf{x}_1, \dots, \mathbf{x}_i, \dots, \mathbf{x}_{\frac{H}{p} \times \frac{W}{p} \times C \times T} \}$. The image patches are grouped into windows of size (6, 8, 8), where each window spans 6 temporal frames, and for every frame contains an 8×8 grid of patches. This grouping aggregates fine-grained patches into coarser spatiotemporal units. For these windows, a binary mask variable $\mathbf{M} \in \{0,1\}^{\frac{H}{p} \times \frac{W}{p} \times C \times T}$, is generated, assigning to each window a value of 0 (masked) or 1 (unmasked). This provides the input data for the reconstruction task of the model.

To enhance learning, the RingMo-Lite PIMask approach is applied, allowing 25\% of the patches within the masked windows to remain unmasked. Patches can be filtered using element-wise multiplication $x_{visible} = \mathbf{M} \odot \mathbf{x}$. The mask is broadcast across time to prevent information being leaked between timestamps. The model uses a combination of patch embedding, relative position biases, and window-based attention structures to effectively encode spatial information without relying on positional embeddings. This is possible due to the Swin transformer that parses all patches instead of only the unmasked patches. Therefore, all processes of patch embedding can be reverted. Despite the higher cost, the benefits outweigh the additional computations to shift window attention. \\

\begin{figure*}[htbp]
    \centering
    \makebox[\textwidth][c]{\includegraphics[width=0.6\linewidth]{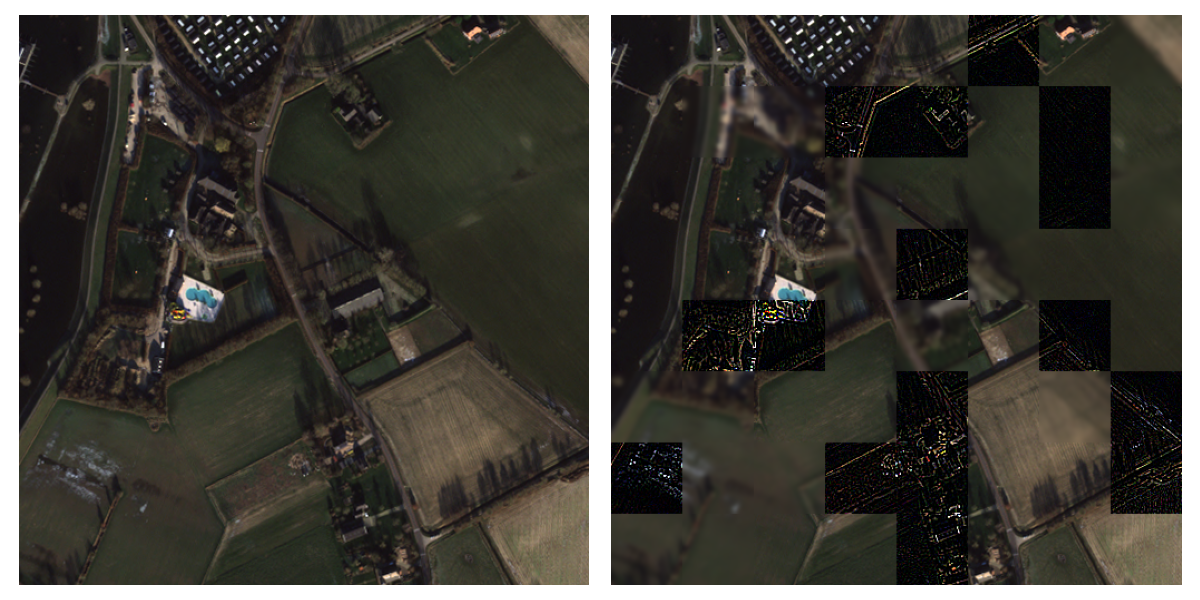}}
    \caption[Example of applying low- and high-pass filters to a pre-training sample.]{Example of applying low- and high-pass filters to a pre-training sample which is shown on the left. The right shows the output after applying the filters, with the window size doubled to enhance visibility for illustration purposes.}
    \label{fig:lowhighpass}
\end{figure*}

A key feature of RingMo-Lite is its frequency-domain training approach, which replaces 50\% of the patches during training with versions filtered by a low-pass or high-pass Fourier transform. This emphasizes specific frequency characteristics that help improve learning and generalization. Although this acts as a data augmentation step, it is applied at the window level rather than the image level. Figure~\ref{fig:lowhighpass} illustrates the effect of the low- and high-pass filters applied across both temporal and spectral patch windows.

\subsubsection{Architecture design}
The architecture follows the Swin Transformer \cite{liu2021swin}, using four patch merging stages to build a hierarchical representation of low- and high-level features. These stages are extended with a CNN branch, as shown in Figure \ref{fig:proposed_architecture}. The advantage of this hybrid structure is that CNNs excel at high-frequency content and \glspl{VIT} at low-frequency content. The CNN-\gls{VIT} blocks between patch merges are repeated 2, 2, 6, and 2 times, respectively, to capture increasingly complex features, with the third stage focusing on global representations. By combining CNNs and \glspl{VIT}, the model maintains low computational complexity while enabling rich feature extraction using a shallower network. The resulting base architecture for the \gls{FM} is Swin-T. To highlight the architecture’s capabilities, the \gls{FM} is trained from scratch without using any pre-trained ImageNet weights, relying solely on \gls{RS} data. \\

The proposed architecture combines two parallel branches to capture both local and global features. The high-frequency branch uses a lightweight CNN to extract local features with small 2D kernels, which operate separately across the spectral and temporal dimensions. This design captures high-frequency variations unique to individual bands or timestamps, while reducing dimensionality for computational efficiency. The low-frequency branch employs the vanilla Swin Transformer to extract global features using window-based self-attention. This mechanism limits attention computation from all patches to within local windows, significantly reducing complexity from $O(N^2)$ to $O(N \cdot M^2)$, where $N$ is the total number of patches and $M$ the number of patches per window. Despite this constraint, the shifting window strategy enables the model to approximate global attention. Additionally, attention is extended to the temporal dimension by applying it both within spatial windows and across corresponding temporal positions. Feature maps from both branches are fused via element-wise multiplication, allowing the model to integrate fine-grained and contextual information efficiently. 


\subsubsection{Reconstruction loss}
The loss function is defined in Equation \ref{eq:loss} for the 5D input tensor (batch, time, channel, height, width). The data tensor has a 5-dimensional shape: batch size (B), temporal size (T), channels (C), height (H), and width (W). The loss function calculates the difference between the input $x$ and the reconstructed image $\hat{x}$, where $m$ represents the number of unmasked pixels. The model learns spatial, spectral, and temporal relations by computing both spatial-to-spatial and spectral-to-spectral loss across the time series, adopted from \cite{hong2023spectralgpt}. The spatial-to-spatial loss first sums the spectral bands before calculating the difference, whereas the spectral-to-spectral loss is calculated based on the difference between the ground truth and the reconstructed patches. Combining these two losses enforces a rich embedding for accurate reconstruction in all spectral channels. 

\begin{equation}
\label{eq:loss}
\begin{aligned}
\mathcal{L} &= \mathcal{L}_{\text{spatial}} + \mathcal{L}_{\text{spectral}} \\
&= \frac{1}{m} \sum_{b,t,c,h,w} mask_{b,t,c,h,w}  \cdot (x_{b,t,c,h,w} - \hat{x}_{b,t,c,h,w})^2 + \\ 
&\quad \frac{c}{m} \sum_{b,t,h,w} (\sum_{c} x_{b,t,c,h,w} \cdot mask_{b,t,c,h,w} - \\&     \sum_{c} \hat{x}_{b,t,c,h,w} \cdot mask_{b,t,c,h,w})^2 \\
\end{aligned}
\end{equation}

\section{Results}
\label{sec:results}

This section covers the pre-training of the \gls{FM} and the fine-tuning of the \gls{FM} to several downstream tasks. The downstream tasks validate and quantify the pre-trained \gls{FM} performance by comparing it to existing performance measures from research on \gls{RS} models. The downstream tasks cover segmentation, land-use classification, and change detection tasks on the datasets: Vegetation Monitoring, RESISC-45 \cite{resisc45}, UC-Merced \cite{ucmerced}, Potsdam \cite{potsdam_dataset}, and Levir-CD \cite{levircd}. The experiments and results are analyzed to evaluate and discuss the performance and capabilities of the trained \gls{FM}. 

\subsection{Downstream Tasks} \label{downstream_tasks}

This subsection covers the evaluation for assessing the performance of the \gls{FM}. The model is fine-tuned and tested on a vegetation monitoring dataset from the Netherlands, as well as on several widely used global remote sensing datasets. For the Dutch dataset, an existing study already trained the SegFormer model \cite{xie2021segformer}, which will serve as a baseline for comparison. To ensure a fair evaluation of the encoder representations, the \gls{FM} uses the same decoder head as SegFormer. This decoder consists of three \gls{MLP} layers that process multi-scale features from different encoder stages.

In addition to the Dutch dataset, the \gls{FM} is also fine-tuned and tested on commonly used remote sensing datasets. While no other models are currently trained on a high-resolution Dutch benchmark, these global datasets allow for comparison with \gls{SOTA} models covered in the literature. This provides valuable insight into the generalization capability of the \gls{FM}. Although the training data is geographically limited to the Netherlands, the evaluation covers diverse landscapes, demonstrating the flexibility and robustness expected of a foundation model.

The evaluation includes tasks such as semantic segmentation, land-use classification, and change detection, using the following datasets: Vegetation Monitoring, RESISC-45, UC-Merced, Potsdam, and Levir-CD. As noted earlier, for segmentation tasks, the SegFormer decoder head is used. For classification, a simplified head is used based on the SegFormer head, ending in a linear classification layer. For change detection, the FDAF-head from RSBuilding is used, which is designed to highlight temporal differences through feature aggregation \cite{wang2024rsbuilding}. Across all tasks, decoder heads are intentionally kept lightweight to emphasize the contribution of the encoder representations.

\begin{table*}[h!]
\centering
\caption[Performance measures for all vegetation classes for the Foundation Model compared to benchmarks.]{Precision and recall scores for all vegetation classes with the improvement of the Foundation Model compared to benchmarks. Best F1 per class in bold.}
\label{tab:pocmvpimprovements}
\begin{adjustbox}{width=\textwidth}
\begin{tabular}{lccc|ccc|ccc}
\toprule
& \multicolumn{3}{c}{Precision (\%)} 
& \multicolumn{3}{c}{Recall (\%)} 
& \multicolumn{3}{c}{F1 (\%)} \\

 & SegFormer 1.2m  & FM 1.2m &  
 & SegFormer 1.2m & FM 1.2m &  
 & SegFormer 1.2m & FM 1.2m & Improvement \\

\midrule
Water        & 96 & \textbf{97} &  & 95 & \textbf{97} &  & 95.5 & \textbf{97.0} & +1.5 \\
Hard Surface & \textbf{81} & 75 &  & 74 & \textbf{76} &  & \textbf{77.3} & 75.5 & -1.8 \\
Grass        & \textbf{94} & 93 &  & \textbf{94} & \textbf{94} &  & \textbf{94.0} & 93.5 & -0.5 \\
Reed         & 61 & \textbf{65} &  & 51 & \textbf{72} &  & 55.6 & \textbf{68.3} & +12.7 \\
Woods        & 72 & \textbf{79} &  & 76 & \textbf{80} &  & 73.9 & \textbf{79.5} & +5.6 \\
Thicket      & 40 & \textbf{55} &  & 41 & \textbf{57} &  & 40.5 & \textbf{56.0} & +15.5 \\

\midrule
 & SegFormer 0.3m  & FM 0.3m &  
 & SegFormer 0.3m & FM 0.3m &  
 & SegFormer 0.3m & FM 0.3m & Improvement \\

\midrule
Water        & 97 & \textbf{98} &  & 96 & \textbf{98} &  & 96.5 & \textbf{98.0} & +1.5 \\
Hard Surface & 74 & \textbf{83} &  & \textbf{81} & 80 &  & 77.3 & \textbf{81.5} & +4.2 \\
Grass        & \textbf{95} & 92 &  & 93 & \textbf{94} &  & \textbf{94.0} & 93.0 & -1.0 \\
Reed         & 58 & \textbf{75} &  & 65 & \textbf{67} &  & 61.3 & \textbf{70.8} & +9.5 \\
Woods        & \textbf{78} & 75 &  & \textbf{77} & \textbf{77} &  & \textbf{77.5} & 76.0 & -1.5 \\
Thicket      & 52 & \textbf{73} &  & 66 & \textbf{74} &  & 58.2 & \textbf{73.5} & +15.3 \\

\bottomrule
\end{tabular}
\end{adjustbox}
\end{table*}

\subsection{Evaluation on Vegetation Monitoring Dataset} \label{vegetation_monitoring}

\begin{figure}[H]
    \centering
    \includegraphics[width=.9\linewidth]{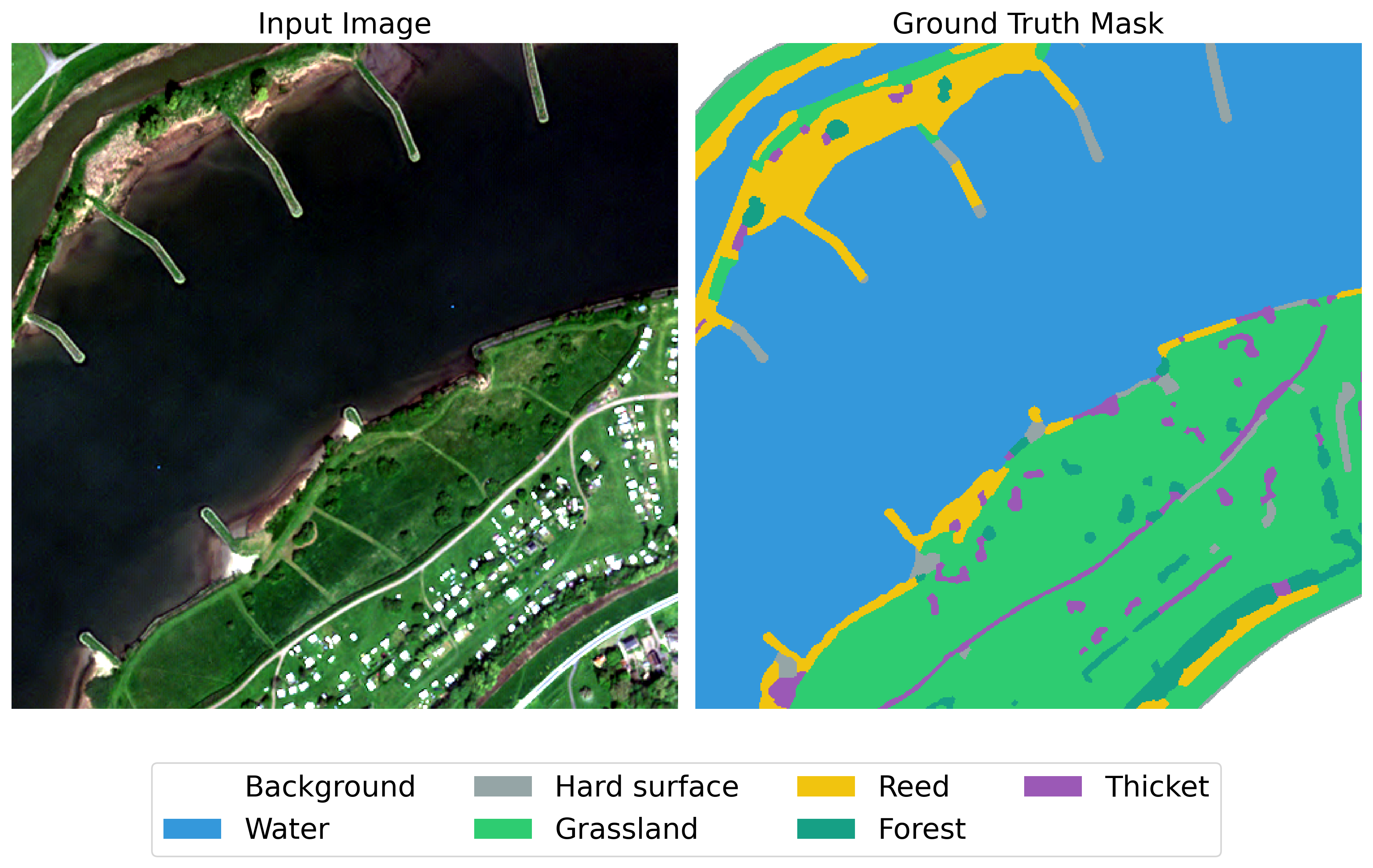}
    \caption[A dataset sample from the vegetation monitoring dataset.]{A dataset sample from the vegetation monitoring dataset containing an input image and the ground truth mask.}
    \label{fig:vegetation_illustration}
\end{figure}

The Vegetation Monitoring dataset maps the vegetation surrounding the large rivers of the Netherlands and validates the \gls{FM} on a segmentation task. The 27 types of vegetation are reduced to 6 umbrella classes, namely, water (31\%), hard surface (5\%), grassland (54\%), reed (4\%), forest (4\%), and thicket (2\%). Even after merging classes, the dataset remains highly imbalanced due to water and grassland covering most of the selected area. Figure \ref{fig:vegetation_illustration} shows a sample from the dataset with the accompanying ground truth mask. \\

Table \ref{tab:pocmvpimprovements} reports precision and recall values, showing improved performance using the temporal datasets with the \gls{FM} compared to the non-temporal SegFormer model. It shows that the model accurately generalizes simpler classes like Water but struggles with more challenging classes such as Reed, Forest, and Thicket, which have similar appearances and fewer samples, making it harder for the model to distinguish them accurately.

\subsection{Evaluation on Benchmark Datasets} \label{generaldownstreamtask}

The general benchmarking datasets consist of different locations around the world that differ noticeably from the landscape characteristics of the Netherlands and, therefore, are more of a challenge than the vegetation monitoring task. The datasets covered in this section are RESISC-45, UC-Merced, Potsdam, and Levir-CD, which address segmentation, land-use classification, and change detection tasks. The performance has been compared to existing models using a table showing the model's name, general pre-training images, remote sensing pre-training images, parameter size, and performance measures. The image count is not directly comparable due to different image sizes and \gls{GSD}. However, these indicate the scale on which the model is trained. It is also noticeable that some models use pre-trained weights of ImageNet, and others start from scratch using only \gls{RS} data. 

Not all datasets could be directly processed, as some lacked train/val/test splits, and others consisted of varying image sizes or channels. The RESISC-45 and UC-Merced datasets did not provide a split. Thus, the data is split according to \cite{mendieta2023towards}, which uses the split provided by Bias Variance Labs \cite{aitlasarena}. The Potsdam dataset is split according to MMSegmentation \cite{mmsegmentation2020}, which is used in \cite{muhtar2023cmid}. The Potsdam dataset is tiled from 6000x6000px scenes into 512×512px patches with 256px stride in two spatial dimensions.

\begin{table*}
\centering
\caption[Top-1 Accuracy performance on the RESISC-45 dataset.]{Top-1 Accuracy results for the Foundation Model on the RESISC-45 dataset for both a frozen and unfrozen (fine-tuned) decoder and a 10\% and 20\% partition of the dataset sorted in ascending order. The first section lists benchmark performances of commonly used backbone models without any fine-tuning on \gls{RS} data. The results presented here are taken from the original works and have not been validated independently. *The performance measures are as reported by \cite{xiong2024neural}.}
\label{tab:resisc45}
\begin{adjustbox}{width=.95\textwidth}
\begin{tabular}{llllcc}
\toprule
Model & General pre-training & Remote sensing pre-training & Parameters & \multicolumn{2}{c}{Top-1 Accuracy} \\
\cmidrule(lr){5-6}
 &  & & & Frozen & Fine-tuned $\blacktriangle$ \\
\midrule
MAE \cite{he2022masked} & 14.0M (ImageNet-21k) & - & 307M (ViT-L) & 88.90 & 93.30 \\ 
\midrule
DOFA \cite{xiong2024neural} & 14.0M (ImageNet-21k) & 0.50M (Diverse) & 86M (ViT-B)  & 91.30  & 93.80 \\ 
SatMAE* \cite{cong2022satmae} & 14.0M (ImageNet-21k) & 0.70M (fMoW-Sentinel) & 307M (ViT-L)  & 88.30  & 94.80 \\ 
ConvMAE* \cite{gao2022convmae} & 1.3M (ImageNet-1k) & - & 195M (ConvVit-L) & 81.20  & 95.00 \\ 
Scale-MAE* \cite{reed2023scale} & 14.0M (ImageNet-21k) & 0.40M (FMoW-RGB) & 307M (ViT-L) & 89.60 & 95.70 \\ 
DOFA \cite{xiong2024neural} & 14.0M (ImageNet-21k) & 0.50M (Diverse) & 307M (ViT-L) & 91.90& 96.10 \\ 
SatMAE++ \cite{noman2024rethinking} & 14.0M (ImageNet-21k) & 0.70M (fMoW-Sentinel) & 307M (ViT-L) & & 97.50  \\ 
SatlasNet \cite{bastani2023satlaspretrain} & - & 0.90M (SatlasPretrain) & 88M (Swin-B) &  & 98.00 \\
\midrule
Foundation Model & - & 0.28M & 30M (Swin-T)  & 91.27 & 95.59 \\
\midrule
 &  & &  & \multicolumn{2}{c}{Top-1 Accuracy} \\
\cmidrule(lr){5-6}
  &  & & &10\% & 20\%\\
\midrule
RingMo-Lite \cite{bastani2023satlaspretrain} & - & 0.15M & 30M (Swin-T) & 89.85 & 93.25 \\
\midrule
Foundation Model & - & 0.28M & 32M (Swin-T) & 82.71 & 88.81 \\
\bottomrule
\end{tabular}
\end{adjustbox}

\end{table*}

\begin{table*}
\centering
\caption[Top-1 Accuracy results for the UC-Merced dataset.]{Top-1 Accuracy results for the Foundation Model on the UC-Merced dataset compared to other research sorted ascending. The first section lists baseline metrics of common backbone models without fine-tuning on \gls{RS} data, which can be referred to as a baseline. The second section shows metrics of fine-tuned models, demonstrating slight improvements in both performance and parameters. The results presented here are taken from the original works and have not been independently validated.}
\label{tab:ucmerced}
\begin{adjustbox}{width=.95\textwidth}
\begin{tabular}{llllc}
\toprule
Model & General pre-training & Remote sensing pre-training & Parameters & Top-1 Accuracy $\blacktriangle$ \\
\midrule
ViT \cite{mendieta2023towards} & 14.0M (ImageNet-21k) & - & 307M (ViT-L) & 93.10  \\
ResNet \cite{mendieta2023towards} & 1.3M (ImageNet-1k) & - & 26M (ResNet-50) & 98.80  \\
Swin \cite{mendieta2023towards} & 14.0M (ImageNet-21k) & - & 88M (Swin-B) & 99.00  \\
\midrule
SatMAE \cite{noman2024rethinking} & 14.0M (ImageNet-21k) & 0.70M (fMoW-Sentinel) & 307M (ViT-L) & 94.10 \\ 
SeCo \cite{manas2021seasonal} & 1.3M (ImageNet-1k) & 1.00M (SeCo) & 26M (ResNet-50) & 97.10  \\
SatMAE++ \cite{noman2024rethinking} & 14.0M (ImageNet-21k) & 0.70M (fMoW-Sentinel) & 307M (ViT-L) & 97.60 \\
SatlasNet \cite{bastani2023satlaspretrain} & 14.0M (ImageNet-21k) & 0.90M (SatlasPretrain) & 88M (Swin-B)  & 99.00 \\
GFM \cite{mendieta2023towards} & -  & 0.60M (GeoPile) & 88M (Swin-B) & 99.00 \\
RingMo-Lite \cite{wang2023ringmo} & - & 0.15M & 28M (Swin-T) & 99.10 \\
CMID \cite{muhtar2023cmid} & 14.0M (ImageNet-21k) & 1.00M (MillionAID) & 31M (Swin-B) & 99.48 \\
\midrule
Foundation Model & - & 0.28M & 32M (Swin-T) & 98.10 \\
\bottomrule
\end{tabular}
\end{adjustbox}

\end{table*}

\begin{table*}
\centering
\caption[Performance measures for the Potsdam dataset.]{Top-1 Accuracy, F1, and mean Intersection over Union (mIoU) results for the Foundation Model on the Potsdam dataset compared to other research sorted ascending on Top-1 Accuracy. The results presented here are taken from the original works and have not been independently validated. }
\label{tab:potsdam}
\begin{adjustbox}{width=.95\textwidth}
\begin{tabular}{llllccc}
\toprule
Model  & General pre-training & Remote sensing pre-training  & Parameters & mIoU & F1 & Top-1 Accuracy $\blacktriangle$ \\
\midrule
Hybrid U-net \cite{liu2023unet} & 1.3M (ImageNet-1k) & - & 11M (ResNet18) & 86.68 & 92.50 & \\
RingMo-Lite \cite{wang2023ringmo} & - & 0.15M & 42M (Swin-T) & & & 91.15 \\ 
Cross-scale MAE \cite{tang2024cross} & 14.0M (ImageNet-21k) & 1.00M (fMoW) & 307M (ViT-L) & 76.17 & & \\ 
CSPT (UperNet) \cite{zhang2022consecutive} & 1.3M (ImageNet-1k) & 1.00M (MillionAID) & 86M (ViT-B) & 78.80 & & \\
CMID \cite{muhtar2023cmid} & 14.0M (ImageNet-21k) & 1.00M (MillionAID) & 31M (Swin-B) & 85.17 & 91.83 & 91.77 \\
CMID \cite{muhtar2023cmid} & 14.0M (ImageNet-21k) & 1.00M (MillionAID) & 24M (ResNet-50) & 87.04 & 92.81 & 92.74 \\
BFM (UperNet) \cite{cha2023billion} & 14.0M (ImageNet-21k) & 1.00M (MillionAID) & 86M (ViT-B) & & 90.87 & 92.17 \\
BFM (UperNet) \cite{cha2023billion} & 14.0M (ImageNet-21k) & 1.00M (MillionAID) & 2042M (ViT-G) & & 92.12 & 92.58 \\
\midrule
Foundation Model & - & 0.28M & 31M (Swin-T) & 72.87 & 87.48 & 87.94 \\ 
\bottomrule
\end{tabular}
\end{adjustbox}

\end{table*}

\begin{table*}
\centering
\caption[Performance measures for the LEVIR-CD dataset.]{Top-1 Accuracy, F1-score, and mean Intersection over Union results for the Foundation Model on the LEVIR-CD dataset compared to other research. The results presented here are taken from the original works and have not been independently validated.}
\label{table:levircd}
\begin{adjustbox}{width=.95\textwidth}
\begin{tabular}{llllccc}
\toprule
Model & General pre-training & Remote sensing pre-training & Parameters & mIoU & F1  & Top-1 Accuracy \\ 
\midrule
RingMo-lite \cite{wang2023ringmo} & - & 0.15M & 54M (Swin-T) & 84.44 & 91.56 & 99.15 \\ 
RSBuilding \cite{wang2024rsbuilding} & 14M (ImageNet-21k) & 0.25M & 30M (Swin-T) & 85.06 & 91.93 &  \\ 
\midrule
Foundation model & - & 0.28M & 33M (Swin-T) & 78.74 & 88.10 & 98.83 \\ 
\bottomrule
\end{tabular}
\end{adjustbox}
\end{table*}

Tables \ref{tab:resisc45} and \ref{tab:ucmerced} show the results of the different models on the RESISC-45 and UC-Merced datasets, respectively. The first section of rows contains a benchmark to compare fine-tuned models to a commonly used pre-trained model that has not been fine-tuned on \gls{RS} data. The performance measure used is Top-1 Accuracy, which is extended for RESISC-45 to a frozen encoder and a 10\% and 20\% partition of the provided dataset. It shows that the trained foundation model is competitive to \gls{SOTA} models even though it is being trained on a small, less diverse, and not globally targeted dataset. Tables \ref{tab:potsdam} and \ref{table:levircd} show the results for the Potsdam and LEVIR-CD datasets, respectively. These datasets use the measures, mIoU, F1, and Top-1 and show that the \gls{FM} does not outperform the \gls{SOTA} models but, given the small training dataset and small parameter size, offers a strong foundation for further development. It should be noted that these other models were developed specifically for different types of datasets, whereas this \gls{FM} is designed for tasks within the Netherlands using data collected in the Netherlands. As a result, the overlap between the data used during pre-training and the datasets used for downstream tasks is relatively limited. 

\subsection{Pre-training Foundation Model}
The model is pre-trained using 8 NVIDIA V100 Tensors Cores, each providing 16GB of memory, requiring 3 days of training until sufficient stabilization of the reconstruction loss on the validation set after 150 epochs. The model is trained on 280,000 patches of 1.2m \gls{GSD} after merging the training and validation set, resulting in roughly 100,000$km^2$ of satellite imagery of the Netherlands spread over a year.

\section{Conclusion}
\label{sec:conclusion}
This paper introduced a compact foundation model tailored for remote sensing tasks in the Netherlands. By leveraging high-resolution temporal satellite data and a lightweight Swin-T architecture, the model achieved competitive performance while significantly reducing resource requirements. Despite its small size, the foundation model demonstrated strong generalization and flexibility, outperforming or matching larger state-of-the-art models on several tasks. Key design choices such as the use of seasonal imagery and the integration of temporal context, contributed to its robustness and practical relevance. 

The developed foundation model provides the foundation for future developments for a model focused on the Netherlands. Future work may investigate larger Swin architectures, multi-modal inputs, and improved fine-tuning techniques.

\bibliographystyle{IEEEtran} 
\bibliography{references}



\end{document}